\documentclass[letterpaper]{article} % DO NOT CHANGE THIS
\pdfoutput=1
\usepackage{aaai23}  % DO NOT CHANGE THIS
\usepackage{times}  % DO NOT CHANGE THIS
\usepackage{helvet}  % DO NOT CHANGE THIS
\usepackage{courier}  % DO NOT CHANGE THIS
\usepackage[hyphens]{url}  % DO NOT CHANGE THIS
\usepackage{graphicx} % DO NOT CHANGE THIS
\usepackage{natbib}  % DO NOT CHANGE THIS AND DO NOT ADD ANY OPTIONS TO IT
\usepackage{caption} % DO NOT CHANGE THIS AND DO NOT ADD ANY OPTIONS TO IT
\usepackage{algorithm}
\usepackage{algorithmic}

\usepackage{newfloat}
\usepackage{listings}

\usepackage{amsmath}
\usepackage{multirow}

\DeclareCaptionStyle{ruled}{labelfont=normalfont,labelsep=colon,strut=off} % DO NOT CHANGE THIS
\floatstyle{ruled}
\newfloat{listing}{tb}{lst}{}
\floatname{listing}{Listing}
\title{Stroke Extraction of Chinese Character Based on Deep \\Structure Deformable Image Registration}
\author{
    Meng Li\textsuperscript{\rm1}, 
    Yahan Yu\textsuperscript{\rm1,\rm2},
    Yi Yang\textsuperscript{\rm1},
    Guanghao Ren\textsuperscript{\rm1},
    Jian Wang\textsuperscript{\rm1}\thanks{Corresponding authors.}
}
\affiliations{
    \textsuperscript{\rm 1}Institute of Automation, Chinese Academy of Sciences\\
    \textsuperscript{\rm 2}School of Artificial Intelligence, University of Chinese Academy of Sciences\\
    \{meng\_li, yuyahan2020, yangyi, guanghao.ren, jian.wang\}@ia.ac.cn
}

\usepackage{bibentry}
\begin{document}

\maketitle

\begin{abstract}
Stroke extraction of Chinese characters plays an important role in the field of character recognition and generation. 
The most existing character stroke extraction methods focus on image morphological features. 
These methods usually lead to errors of cross strokes extraction and stroke matching due to rarely using stroke semantics and prior information. 
In this paper, we propose a deep learning-based character stroke extraction method that takes semantic features and prior information of strokes into consideration. 
This method consists of three parts: image registration-based stroke registration that establishes the rough registration of the reference strokes and the target as prior information; 
image semantic segmentation-based stroke segmentation that preliminarily separates target strokes into seven categories; 
and high-precision extraction of single strokes. In the stroke registration, we propose a structure deformable image 
registration network to achieve structure-deformable transformation while maintaining the stable morphology of single 
strokes for character images with complex structures. In order to verify the effectiveness of the method, 
we construct two datasets respectively for calligraphy characters and regular handwriting characters. 
The experimental results show that our method strongly outperforms the baselines. Code is available at https://github.com/MengLi-l1/StrokeExtraction.
\end{abstract}

\section{Introduction}

Stroke extraction of Chinese characters refers to extracting every single stroke of the characters based on the matching with 
the templates consisting of standard ordered strokes. Stroke extraction is important for certain research on Chinese characters. 
In the field of character recognition, the experiments in \cite{xiao2017building,zhang2020radical} show that further disassembly of the stroke structure 
of characters can significantly improve the accuracy of character recognition. In other search fields such as evaluation of calligraphy the 
important part of traditional Chinese culture \cite{wang2016evaluation}, and character generation \cite{liu2021fontrl}, the stroke extraction is also of great significance to them.

By analyzing the characteristics of Chinese characters, the difficulties of Chinese character stroke extraction mainly include the following three aspects. 
First, there are more than 7000 Chinese characters commonly used. Most of them have a complex structure. 
Second, the shapes of character strokes are simple and only have structural differences. 
These indistinguishable features make recognition directly of stroke hard for cross strokes within character. 
Third, the unfixed number of strokes in different Chinese characters makes it difficult to build a stroke extraction model.

Existing research on stroke extraction of Chinese characters, including some deep learning-based methods, 
mostly focus on image morphological features of strokes and radicals \cite{kim2018semantic,tseng1992efficient,wang2022query}. 
Although these methods can realize remarkable results, their cores are only morphological analysis, which exists two drawbacks or constraints: 
(1) the prior information of the character stroke is rarely used, and (2) the semantic analysis of the stroke is lacking. 
Rarely using prior information and stroke semantics can lead to errors in the separation of cross strokes and the matching of strokes with the template 
in the stroke extraction of complex characters.

Inspired by the stroke extraction process of humans, we propose an efficient stroke extraction method of 
Chinese characters which not only separates strokes but also establishes the matching with the reference template. 
This method takes semantic features and prior information of strokes into consideration and mainly includes three steps: 
(1) stroke registration based on the Structure Deformable Image Registration Network (SDNet); 
(2) stroke segmentation based on the Image Semantic Segmentation Network (SegNet); 
(3) single stroke high-precision extraction based on the Single Stroke Extraction Network (ExtractNet).

For human cognition, the prior information of Chinese character images refers to the basic knowledge of the position and shape of the strokes. 
To obtain the prior information, we use the image registration to establish a rough mapping relationship between the reference character strokes and the target character. 
The transformed reference strokes based on the mapping relationship are used as prior information of the target stroke positions and shapes. 
The semantic features of the strokes need to be stable during the registration-based transformation for effective prior information. However, 
the existing image registration methods \cite{wang2022query,haskins2020deep,sotiras2013deformable} usually 
cause the Chinese stroke to be severely distorted when characters have complex structures. 
To solve the problem, we propose SDNet using multiple linear mapping planes to replace the native single mapping surface. 
SDNet can maintain the stability of stroke semantic features while transforming deformably stroke structure. Our main contributions are summarized as follows.

\begin{itemize}
\item We propose a novel deep learning-based stroke extraction method, which takes semantic features and prior information of strokes into consideration more 
adequately and achieves significant improvement in the separation of cross strokes and matching of strokes with the template.
\item We propose a structure deformable image registration method, which performs better in the registration of image structure.
\end{itemize}

\section{Related Work}

\subsection{Image Registration}
Image registration is a process of establishing pixel-level correspondences between different images with matched image contents. 
In the past decades, image registration usually extracted and matched feature regions first, such as closed-boundary regions, edges, corners, 
and other shape features \cite{zitova2003image}. By evaluating the transformation model, like elastic model \cite{davatzikos1997spatial,shen2002hammer} 
and discrete model \cite{dalca2016patch,glocker2008dense}, the transformation relationship of these feature regions and entire image is established. 
Later, some researchers began to use deep learning techniques to enhance the feature extraction and matching, 
based on an iterative framework or reinforcement learning \cite{cheng2018deep,haskins2020deep,liao2017artificial}. 
Recently, with the proposal of Spatial Transformer Networks (STN) \cite{jaderberg2015spatial}, the grid sample block with gradient backpropagation has 
facilitated the direct application of deep learning \cite{balakrishnan2019voxelmorph,vos2017end}, which effectively promotes the improvement of the image registration. 
However, the existing deep learning-based image registration methods cannot maintain the local stability while the whole structure is freely transformed. 
It is not suitable for stroke registration of Chinese characters with complex structures.

\subsection{Stroke Extraction for Chinese Character Image}
For most existing stroke extraction methods of Chinese character, analyzing the ambiguous cross region is the core and primary task. 
By detecting the corners caused by the interlaced strokes, \cite{sun2014geometric} disassembled the Kaiti characters into simple strokes for calligraphy robots. 
\cite{he2000stroke} used chained line segments to represent the boundaries of characters and separated interlaced strokes by detecting whether these boundaries are regular. 
To further improve the accuracy of stroke extraction, some template-based methods are proposed \cite{cao2000model,wang2022query}. 
The methods use stroke structure matching to establish the correspondence between sub-strokes and template strokes, 
which is used to merge sub-strokes created by separating strokes with cross region. 
However, the template of the existing template-based methods is not used for previous steps of cross region detection and separation. 
In this case, the reference template information (prior information) is insufficiently used. In addition, character structure matching is mostly based on shape 
analysis and lacks the use of stroke semantic information.

\begin{figure}[t]
    \centering
    \includegraphics[width=0.9\columnwidth]{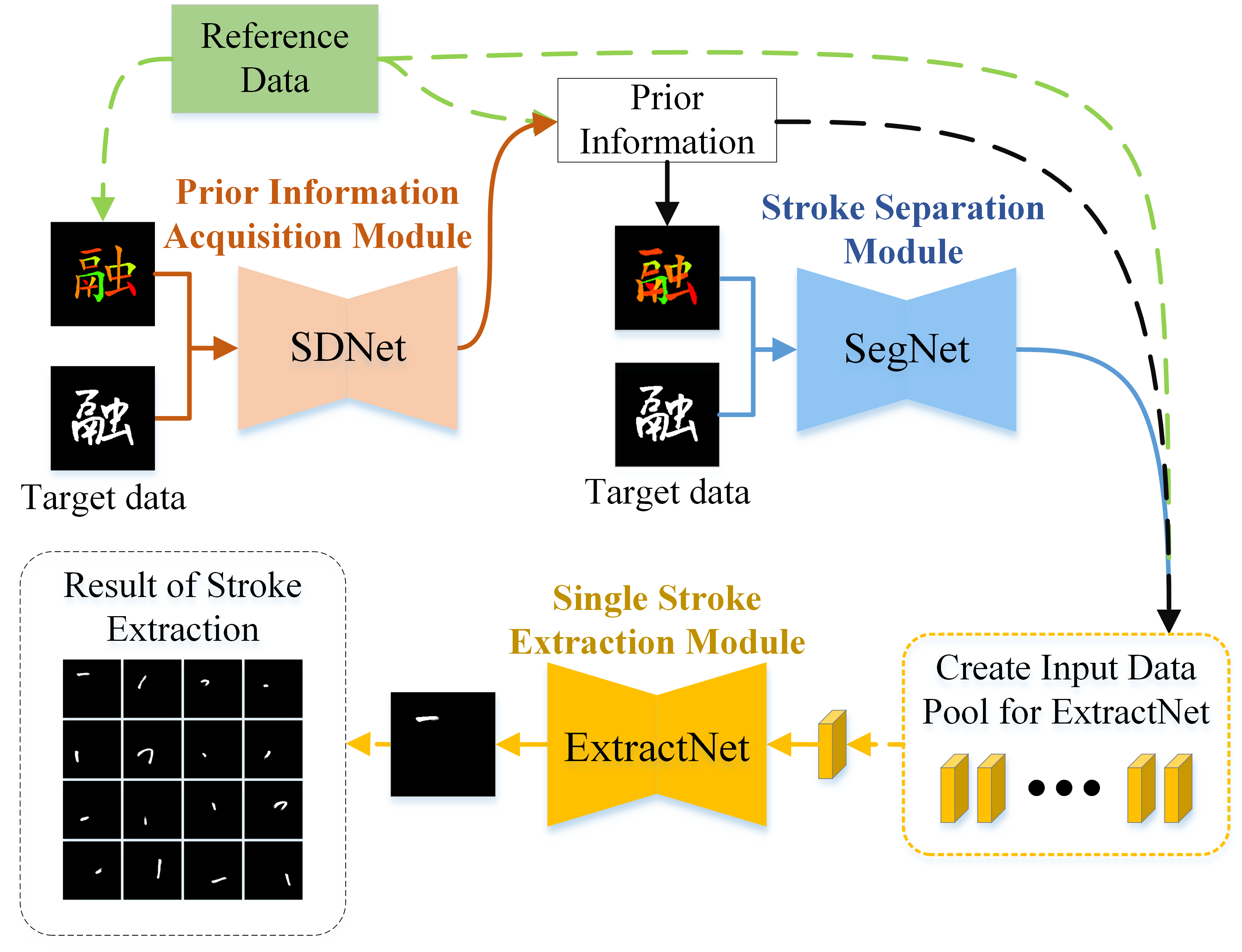} % Reduce the figure size so that it is slightly narrower than the column. Don't use precise values for figure width.This setup will avoid overfull boxes.
    \caption{The pipeline of our method. The main modules are marked by different colors.}
    \label{fig1}
\end{figure}

\section{Method}
The proposed stroke extraction method, shown in Figure 1, mainly includes the following three modules. 
(1) The prior information acquisition module that establishes the registration between the reference stroke and the target through SDNet and uses the transformed reference strokes as prior information. 
(2) The stroke separation module that separates the target character preliminarily by SegNet with the guidance of prior information. 
(3) The single stroke extraction module that uses ExtractNet to extract every single stroke images of the target one by one according to the order of the reference strokes.

\subsection{SDNet for Prior Information}
Due to the complex stroke structure and various writing styles of Chinese characters, the position and shape of the stroke in the same character may be very different. 
It is the reason that it needs the high deformable registration model. However, highly distorted strokes can destroy their own shapes and reduce the validity of prior information, 
which needs the registration model to ensure that the shape of a single stroke is stable before and after transformation. For addressing the problem, we propose a local linear stroke 
spatial transformation method that constrains the transformation of a single stroke to be linear.

\begin{figure*}[h]
    \centering
    \includegraphics[width=0.95\textwidth]{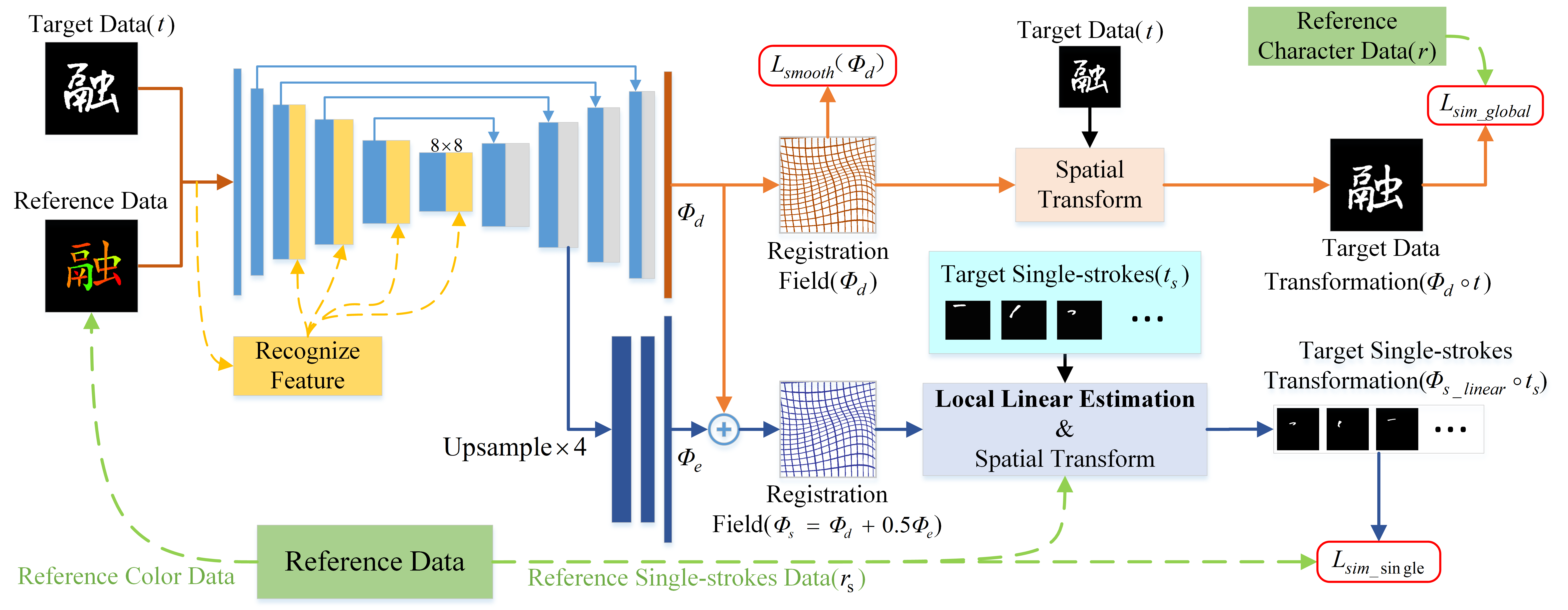} % Reduce the figure size so that it is slightly narrower than the column.
    \caption{The structure of the SDNet.}
    \label{fig2}
\end{figure*}

The SDNet uses UNet as the main frame of the registration network similar to \cite{balakrishnan2019voxelmorph}. 
The main frame convolves down from size $256\times 256$ to $8\times8$ and then convolves up to size $256\time256$ for output. 
To improve the analysis of Chinese character features, we add features of the Chinese character recognition of the input characters in the last four stages of encoder in the UNet.
The Chinese character recognition model is a simple convolution network like VGG \cite{simonyan2014very}. 
The input of SDNet consists of two parts: target character image input as Target Data  $\mathit{t}$ and reference character image marked with different values for different stroke labels input as Reference Data. 
The output of the model is a prediction of the offset coordinate vector for each pixel, which can be easily constructed as a registration field. The structure of The SDNet is shown in Figure 2.

Based on the existing output registration field $\mathit{\Phi_d}$, as shown in Figure 2, we add a branch in the up-convolution process. 
This branch upsamples the data at size $32\time32$ by a factor of 4 and then upsamples again after one convolution to obtain the registration field $\mathit{\Phi_e}$ with the same size as $\mathit{\Phi_d}$. 
$\mathit{\Phi_e}$ is a fine-tuning of the original registration field, and its output weight is only 0.5 of $\mathit{\Phi_d}$. 
The registration field used to calculate the linear transformation is represented as $\mathit{\Phi_s}$ ($\mathit{\Phi_s=\Phi_d+0.5\Phi_e}$). 
Due to learning from a smaller size, $\mathit{\Phi_e}$ is biased towards th e prediction of the overall offset of the single stroke, 
which is suitable for the estimation of local linear transformation. During model training, both $\mathit{\Phi_d}$ and $\mathit{\Phi_s}$ are involved in the loss computation. 
$\mathit{\Phi_s}$ tends to learn the local registration ability of strokes under the constraint, which weakens the learning of global registration. 
Therefore, $\mathit{\Phi_d}$ is used to realize the global registration of character images to make up for the lack of $\mathit{\Phi_s}$.

In actual training, the position and shape of the reference stroke are more stable. 
We use the reference stroke to mark the local region in $\mathit{\Phi_s}$ and calculate the linear transformation estimation of these regions. 
Therefore, during model training, what we actually learn is the transformation from target to reference. 
This operation can reduce the noise caused by errors in the linear estimation part, and improve the stability and efficiency of training. 
During inference, the transformation from reference to target can be obtained by calculating the inverse spatial transformation for every single stroke.

\subsubsection{Linear Estimation of Single Stroke Spatial Transformation}
The existing linear fitting methods cannot be embedded in deep networks to achieve gradient backpropagation. 
Inspired by the Taylor series, we construct a linear estimation method that can be used in deep neural networks:
\begin{equation}
    \begin{split}
        \varPhi_{s\_linear} =& mean(\varPhi_{s\_local})+(\boldsymbol X-P_x) \times \\
        mean(\frac{\partial \varPhi_{s\_local}}{\partial \boldsymbol X})&+ (\boldsymbol Y-P_y) \times mean(\frac{\partial \varPhi_{s\_local}}{\partial \boldsymbol Y}),
    \end{split}
\end{equation}
where $\varPhi_{s\_local}$ represents the local region of $\varPhi_s$ used for linear estimation. 
$\boldsymbol X$ and $\boldsymbol Y$ denote coordinate matrices. 
$P_x$ and $P_y$ denote the coordinates of the centroid of the local region, which can be calculated from the reference stroke. 
$\varPhi_{s\_linear}$ denotes the linear estimation result. Equation 1 is similar to the Least Squares Linear Fitting method while the slope is 
directly estimated as the average of the gradients to simplify the calculation. Due to the strong learning ability of deep learning, 
this simplification will not affect the final estimation result, but it can effectively reduce the computing workload.

\subsubsection{Loss for Training}
The loss of SDNet consists of two parts: the global registration loss $L_{global}$ and the single-stroke registration loss $L_{single\_linear}$. 
$L_{global}$ includes similarity loss $L_{sim\_global}$ and smoothing loss $L_{smooth}$ of $\varPhi_d$. 
$L_{single\_linear}$ is the average of all single stroke similarity losses. 
For the operation of linear transformation estimation, we do not need to calculate the smoothing loss of $\varPhi_s$.

Traditional image registration methods usually use Normalization Mutual Information (NCC) to measure the similarity of two images. 
However, NCC is not suitable for the data with simple shape such as the stroke image. 
To solve this, we build a ContentNet that is trained to auto-encode stroke images with a simple Encoder-Decoder structure. 
The similarity loss of the two stroke images is defined as the Euclidean distance of the encoding results with $l_2$-normalization:
\begin{equation}
    \begin{split}
        S_c(a,b) = Dis[norm_{l_2}(E(a)),norm_{l_2}(E(b))],    
    \end{split}
\end{equation}  
where $Dis$ denotes the calculation of Euclidean distance. $E$ denotes the access function of the encoding result of ContentNet. 

Final loss in train process is defined as:
\begin{align}
    \begin{split}
        L_{sum}(t,r,t_s,r_s,\varPhi_d,\varPhi_s) = \lambda &L_{single\_linear}(t_s,r_s,\varPhi_s)+ \\
        L_{sim\_global}(r,t,\varPhi_d )+\gamma &L_{smooth}(\varPhi_d),   
    \end{split}\\
    \begin{split}
        L_{single\_linear}(t_s,r_s,\varPhi_s )&= \\
        \frac{1}{stroke_{num}}\sum_{i = 1}^{stroke_{num}}&S_c(r^i_s,\varPhi_{s\_linear}^i \circ t^i_s ),
    \end{split}
\end{align}
\begin{align}
    L_{sim\_global}(r,t,\varPhi_d)&=S_c(r,\varPhi_d \circ t),      \\
    L_{smooth}(\varPhi_d)=mean&(\|\frac{\partial \varPhi_d}{\partial \boldsymbol X}\|^2 + \|\frac{\partial \varPhi_d}{\partial \boldsymbol Y}\|^2),    
\end{align}
where $\varPhi_{s\_linear}^i$ is the linear transformation estimation result corresponding to the reference single stroke $r_s^i$ and the target single stroke $t_s^i$. 
Considering that the global registration result will have a greater impact on $\varPhi_s$, we apply a larger weight to $\varPhi_d$, 
especially to $L_{smooth}$ , and set $\lambda $ and $\gamma $ to be 0.5 and 5, respectively, in order to ensure a stable and better registration result of $\varPhi_d$.

\begin{figure}[t]
    \centering
    \includegraphics[width=1.0\columnwidth]{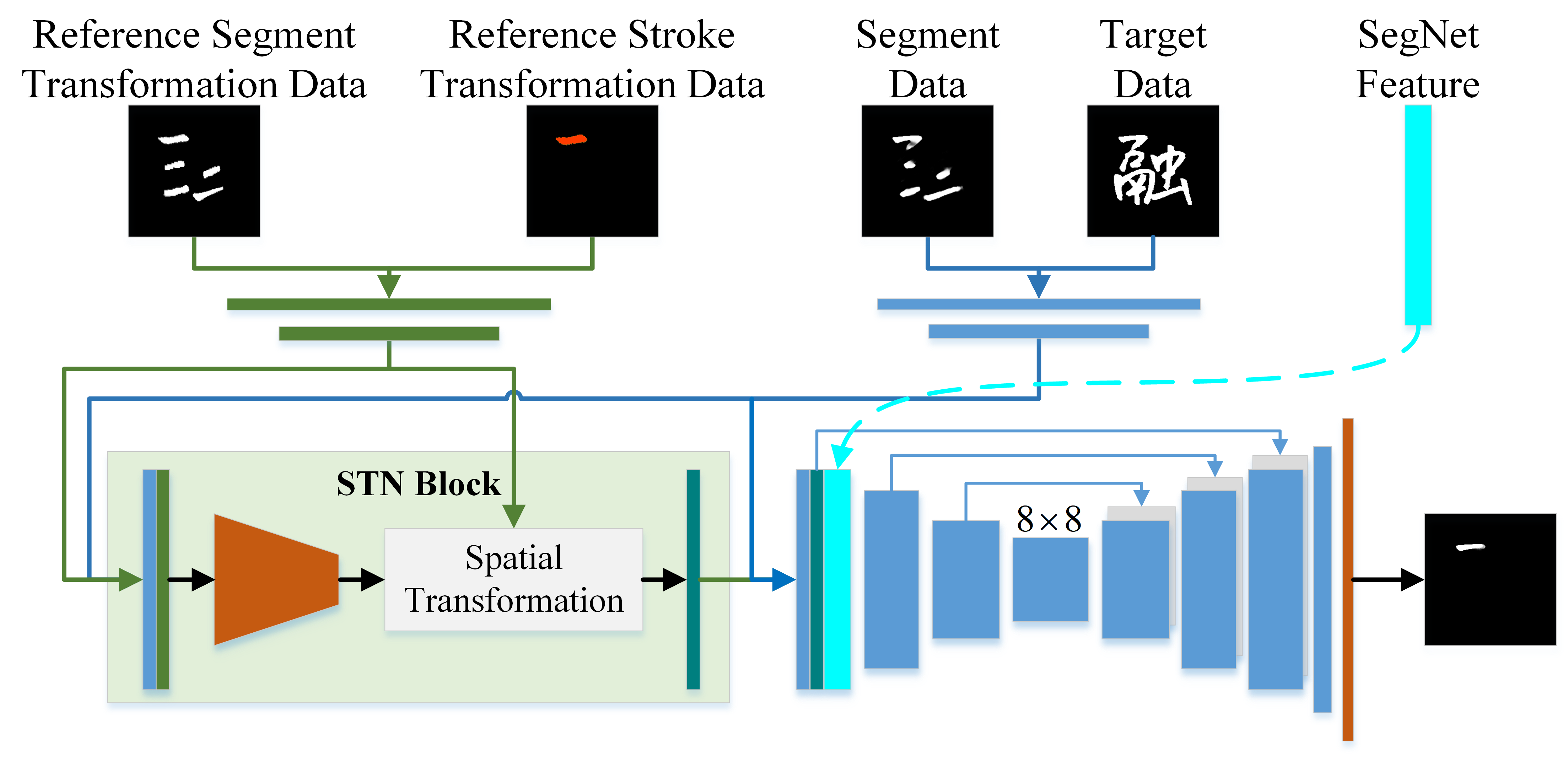} % Reduce the figure size so that it is slightly narrower than the column.
    \caption{The structure of the ExtractNet.}
    \label{fig3}
\end{figure}

\subsection{SegNet for Separating Strokes Roughly}
There are 32 basic strokes for Chinese characters. However, there is a high similarity between these basic strokes, and the number of strokes is seriously unbalanced. 
Therefore, we divide the 32 basic strokes into 7 categories artificially based on the following three rules: 
(1) The number of strokes used in common Chinese characters within the category is as balanced as possible. 
(2) The similarity of stroke shape is as high as possible within the category and as low as possible between the category. 
(3) The probability of stroke crossing within the category is as low as possible.
We use the network architecture adapted from the Deeplabv3 model \cite{chen2017rethinking} as the main frame of the SegNet to segment strokes guided by prior information. 
Considering the cross stroke, we construct the SegNet as a multi-label model. The loss of SegNet is the average of the binary cross-entropy of output and label. 
The input of SegNet consists of two parts: Target Data and Prior Data. Prior Data is composed of reference single strokes that are linearly transformed by SDNet. 
Strokes in Prior Data with different categories are marked with different values. 
In the training process, in order to improve the generalization of SegNet, we apply a random position offset within maximal 5 pixels to every single stroke in Prior Data.

\subsection{ExtractNet for Single Stroke Extraction}
\subsubsection{Input Data}
As shown in Figure 3, ExtractNet has five inputs to provide sufficient prior information and stroke semantic information. 
Table 1 shows the details of these inputs. 
For ExtractNet, Segment Data and Reference Stroke Transformation Data provide the major information required for stroke extraction. 
Considering the possible segmentation errors of SegNet, we add SegNet Feature and Target Data to supplement the information not included in the Segment Data. 
The Reference Stroke Transformation Data can only roughly mark the location and shape of the target stroke. 
For the high-precision stroke extraction work, we add Reference Segment Transformation Data to provide relative positional relationship information 
with Reference Stroke Transformation Data. In this way, through spatial transformation of the STN Block, 
the transformed reference information can be further registered to the target data, which can provide more accurate prior information of stroke position and shape.

Furthermore, within a Chinese character, the size of different strokes may vary greatly, which will weaken the learning of small-sized strokes. 
Therefore, we adaptively scale and crop images of input data and labels to eliminate the size difference between strokes.

\begin{table}[t]
    \centering
    \resizebox{1.0\columnwidth}{!}{
    \begin{tabular}{c|c}
        \hline
        {\bf Name} & {\bf Description}\\
        \hline

        Target Data & Target character image \\

        \hline

        Reference Stroke  & The corresponding transformed single \\
        Transformation Data & reference stroke image in prior information \\

        \hline

        \multirow{3}{*}{Segment Data} & The segmentation result of the category to \\
                                      & which the current stroke belongs in the \\
                                      & SegNet \\
        \hline

                                     &  Divide the transformation reference \\ 
        Reference Segment            &  strokes into seven categories similar to the\\
        Transformation Data          &  SegNet. And select the category data the \\
                                     &  current stroke belongs to \\
        \hline

        \multirow{2}{*}{SegNet Feature} & Feature data at $64\time64$ size of \\
                                        & up-convolution in the SegNet \\
        
        \hline
    \end{tabular}}
    \caption{Detailed descriptions of inputs in the ExtractNet.}
    \label{table1}
    \end{table}

\subsubsection{Structure and Loss}
The structure of ExtractNet is shown in Figure 3, which mainly includes two parts: STN Block and the simple convolution network used to extract strokes. 
In the beginning, we quickly compress the input to 1/4 of the original size by two layers of convolution. 
The STN Block is used to further register the reference information to the target. The output is a single-channel stroke image. 
We use binary cross-entropy to calculate the loss between the output and label.

\section{Experiments}
\subsection{Datasets and Reference Data}
To evaluate our method, we construct two stroke extraction datasets for calligraphy and handwriting, 
which basically cover the main application fields of the stroke extraction.

\begin{itemize}
\item \textbf{The Chinese Calligraphy Character Stroke Extraction Dataset (CCSEDB):} CCSEDB has a total of 5000 character data, consisting of calligraphy characters, 
printed calligraphy characters, and calligraphy characters. We carefully select characters of CCSEDB to maintain a balance between the number of 
strokes and stroke structure. Each record of data in CCSEDB contains a target image and some singlestroke label images of the target image 
arranged in reference stroke order.
\item \textbf{The Regular Handwriting Character Stroke Extraction Dataset (RHSEDB):} We construct RHSEDB referring to \cite{wang2022query} based on the online 
handwriting dataset CASIA-OLHWDB \cite{liu2011casia}, which contains 28,080 pieces of handwriting data written by 17 writers in total. 
The format of each piece of data in RHSEDB is the same as CCSEDB, while the images of writing track of the stroke is normalized to a 
width of 6 pixels (the size of the stroke image is $256\time256$ pixels).
\item \textbf{Reference data:} We construct reference data for CCSEDB and RHSEDB, respectively. 
For CCSEDB, due to the large stroke area, we use character images and the corresponding single stroke images of the Kaiti font as reference data. 
For RHSEDB, due to the thin stroke width, we use the skeleton images and the corresponding single stroke skeleton images of the Kaiti font as reference data, 
which are also normalized to a width of 6 pixels.
\end{itemize}

\subsection{Implementation Detail}
In the experiments, for both CCSEDB and RHSEDB, we use 90\% of the data for training and 10\% for testing. 
The images of training data of SDNet, SegNet, and ExtractNet have resolution of $256\time256$. The training data are binarized, 
except for a few that need to be marked with three-channel value. The three models are trained progressively with a batch size of 8 and for 40, 10, and 20 epochs 
respectively. Their learning rates are initialized to 0.0001 and decrease by a factor of 0.5 every 10, 2, and5 epochs respectively.

\subsection{Stroke Registration Results}
In this task, we build experiments on two representative image registration methods: TpsStn \cite{jaderberg2015spatial} and VoxelMorph \cite{balakrishnan2019voxelmorph}. 
TpsStn uses the idea of STN to realize the TPS registration of two images. 
Due to the fewer control points, this method well maintains local stability but lacks sufficient deformability. 
In order to fully verify the registration effect of TpsStn, we evaluate TpsStn with different numbers of control points of $4\time4$ and $8\time8$ respectively. 
VoxelMorph is a typical image registration method that can theoretically achieve the maximum deformable transformation for predicting the offset of each pixel. 
In order to quantify the effect of the prior information constructed by the registration results. 
For the stroke position, we estimate the qualitative result of position with centroid pixel distance $mDis$ of the single stroke. 
For the stroke shape that is mainly reflected in the size, we estimate the qualitative result of stroke shape with the IOU $mBIou$ of the bounding box of the single stroke.

\begin{align}
    mDis =& \frac{1}{n}\sum_{i=0}^{n}Dis(centroid(rt_s^i),centroid(t_s^i )), \\
    mBI&ou = \frac{1}{n}\sum_{i=0}^{n}IOU(box(rt_s^i),box(t_s^i)),    
\end{align}

\begin{figure}[t]
    \centering
    \includegraphics[width=1.0\columnwidth]{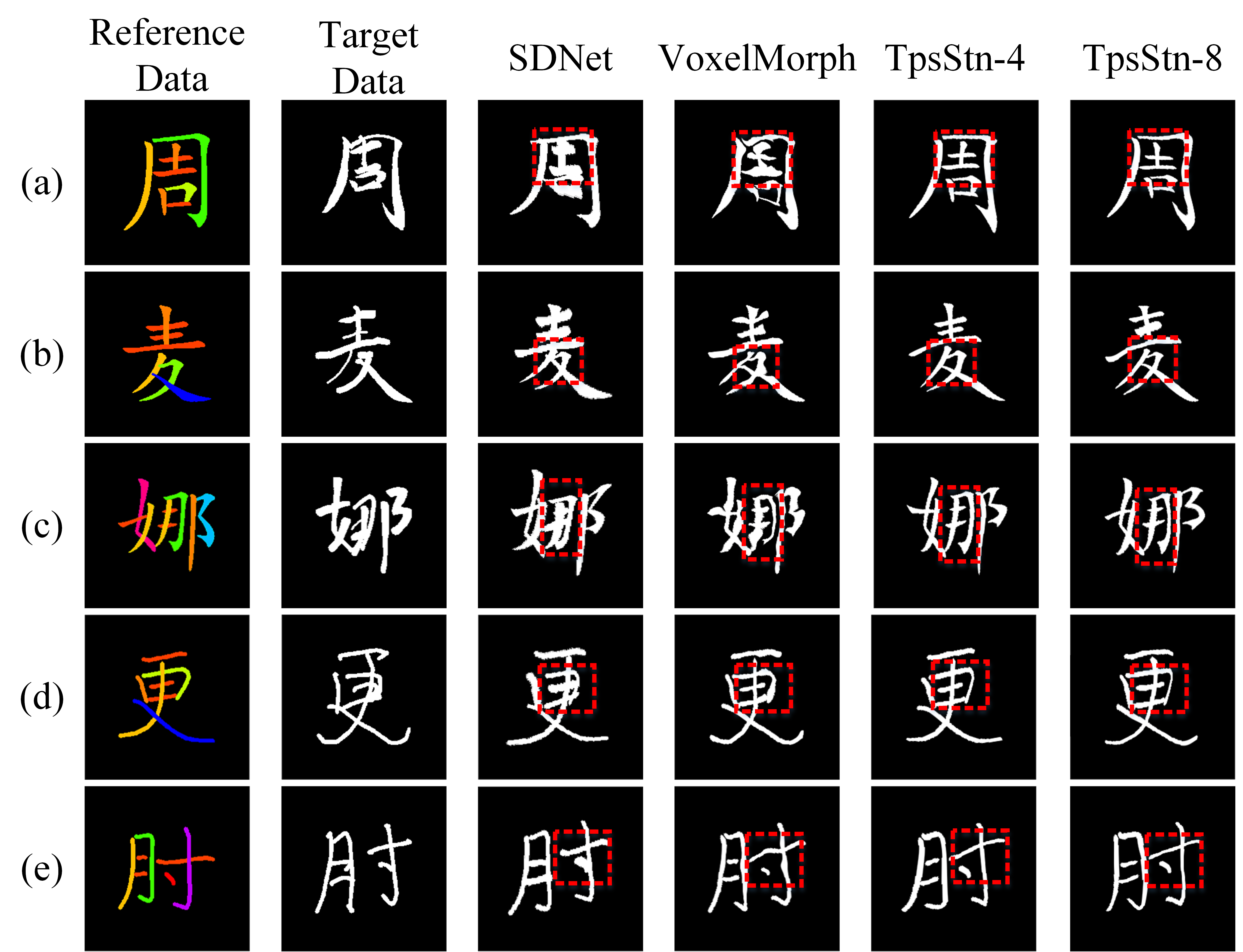} % Reduce the figure size so that it is slightly narrower than the column.
    \caption{Comparison of stroke registration results for our SDNet and baseline models.}
    \label{fig4}
\end{figure}

In Equation 7 and Equation 8, $t_s^i$ denotes the single stroke of the target character. $rt_s^i$ denotes the single transformation reference stroke in prior 
information by SDNet. Dis refers to the Euclidean distance. A smaller mDis and a larger mBIou indicate the higher accurate prior information.

\begin{table}[t]
    \centering
    \resizebox{1.0\columnwidth}{!}{
    \begin{tabular}{c|c|c|c|c}
        \hline   % table title
        \multirow{2}{*}{}      & \multicolumn{2}{|c|}{CCSEDB} &   \multicolumn{2}{c}{RHSEDB} \\
                         \cline{2-5}
                          & $mDis$   &     $mBIou$     &    $mDis$     &     $mBIou$  \\
        \hline
        Reference Data     &   11.7   &     0.365     &    14.763     &    0.277 \\
        \hline
        \textbf{SDNet}	&\textbf{4.469}	&\textbf{0.637}	&\textbf{5.553}	&\textbf{0.614} \\
        \hline
        VoxelMorph	&5.352	&0.609	&7.531	&0.527\\
        \hline
        TpsStn-4	&6.616	&0.533	&7.86	&0.479\\
        \hline
        TpsStn-8	&6.271	&0.551	&7.564	&0.487\\
        \hline
    \end{tabular}}
    \caption{Quantitative results of our SDNet and baseline methods.}
    \label{table2}
\end{table}

As we can see from Table 2, SDNet performs much better than other baseline methods in qualitative results of prior information. 
This means that our method provides more precise prior information to the target character. 
As shown in Figure 4, the SDNet has the best results, which is manifested in higher structural deformability and more stable single-stroke morphology. 
Especially in Figure 4 (d) and (e), this advantage is more prominent when the target structure is quite different from the reference structure. 
We believe that this is mainly due to the registration branch $\varPhi_e$ and linear estimation of single stroke spatial transformation. 
Local linear estimation can construct different transformations for every single reference stroke even for cross stroke while providing linear constraint. 
This is the reason for the structure deformable which is enhanced greatly by the registration branch $\varPhi_e$. 
However, TpsStn and VoxelMorph need to balance the deformation and the smoothness because each of them has only one registration field. 

\begin{figure}[t]
    \centering
    \includegraphics[width=1.0\columnwidth]{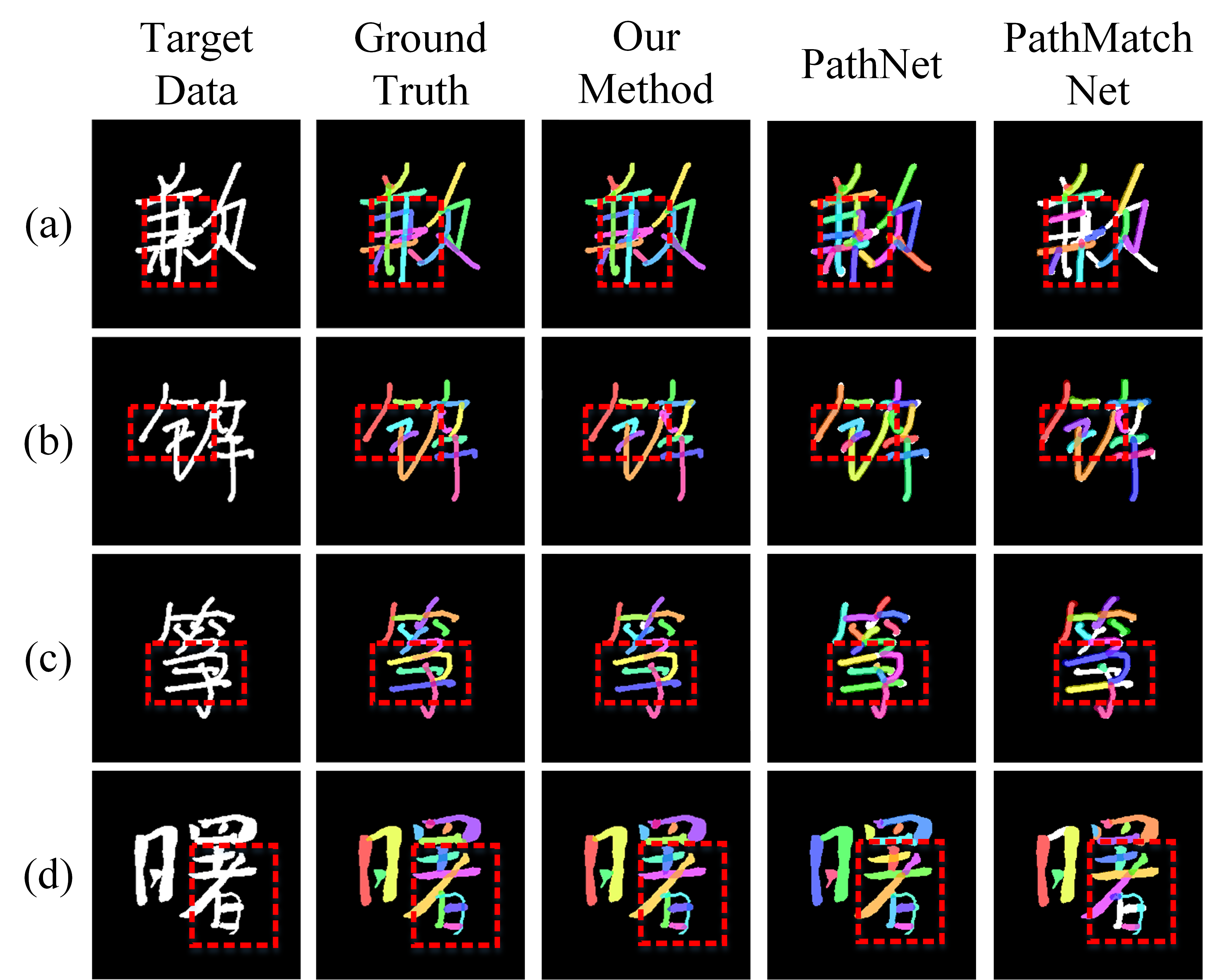} % Reduce the figure size so that it is slightly narrower than the column.
    \caption{Comparison of stroke extraction results of our method and other baseline methods. Different strokes in the results are marked with different colors. 
             The strokes except the PathNet with the same color mean they match the same single reference stroke.}
    \label{fig5}
\end{figure}

\begin{table}[t]
    \centering
    \resizebox{1.0\columnwidth}{!}{
    \begin{tabular}{c|c|c|c|c}
        \hline   % table title
        \multirow{2}{*}{}      & \multicolumn{2}{|c|}{CCSEDB} &   \multicolumn{2}{c}{RHSEDB} \\
                         \cline{2-5}
                          & $mIOU_m$   &     $mIOU_{um}$     &    $mIOU_m$     &     $mIOU_{um}$  \\
        
        \hline
        \textbf{Our Method }	&\textbf{0.924}	&\textbf{0.926}	&\textbf{0.917}	&\textbf{0.93} \\
        \hline
        PathNet	& $\backslash $ 	&0.84	& $\backslash $	&0.789\\
        \hline
        PathMatchNet	&0.692	&0.842	&0.654	&0.815\\
        \hline
    \end{tabular}}
    \caption{$mIOU_m$ and $mIOU_{um}$ of our method and baseline methods.}
    \label{table3}
\end{table}

\subsection{Stroke Extraction Results} 
We find that the morphological analysis method based on deep learning has a great improvement over traditional methods for stroke extraction. 
Therefore we only compare the recent best deep learning-based stroke extraction methods \cite{wang2022query} named Path-MatchNet and \cite{kim2018semantic} named PathNet. 
PathNet separates strokes by predicting the probability that two pixels belong to the same stroke using a deep learning-based image semantic segmentation method. 
Path-MatchNet adds stroke matching on the basis of PathNet to further improve the accuracy of stroke extraction. Referring to \cite{kim2018semantic}, 
we construct two evaluation methods that employ the same evaluation strategy but differ slightly based on whether it needs to match the reference stroke. 
The evaluation considering matching is defined as :
\begin{equation}
    mIOU_m = \sum_{i=0}^{n}IOU(rt_s^i,t_s^i), 
\end{equation}
The evaluation without considering matching is defined as:
\begin{equation}
    mIOU_{um} = \sum_{i=0}^{n}IOU(rt_s^i,maxCross(rt_s^i,t_s)),
\end{equation}

In Equation 9 and Equation 10, $t_s^i$ denotes the single stroke. $t_s$ denotes  all of the single strokes of the target character. 
$rt_s^i$ denotes  the single transformation reference stroke in prior information by SDNet. 
$maxCross$ refers to obtaining the single stroke image of target character with the largest intersection area with $rt_s^i$ in $t_s$.

As shown in Table 3, our method performs much better in $mIOU_{um}$ and $mIOU_m$ than baseline methods. 
As shown in Figure 5, our stroke extraction results have higher precision and are almost 100\% accurate in matching with reference strokes, 
which benefit from the use of prior information and stroke semantic information. As shown in Figure 5 (a) and (d), 
PathNet and PathMatchNet have lower stroke extraction precision in the cross stroke area. 
That is because they only use deep learning for morphological stroke trajectory analysis, and lack analysis of the semantics. 
In the part of stroke matching, simple position and morphological feature similarity calculation in PathMatchNet cannot guarantee the accuracy of matching, 
especially in Chinese characters with a large number of strokes, as shown in Figure 5 (a) and (b).

\subsection{Ablation Study}
To evaluate the effect of prior information and stroke semantic information, we design ablation experiments for SegNet and Extraction. 
\begin{figure}[hb]
    \centering
    \includegraphics[width=0.73\columnwidth]{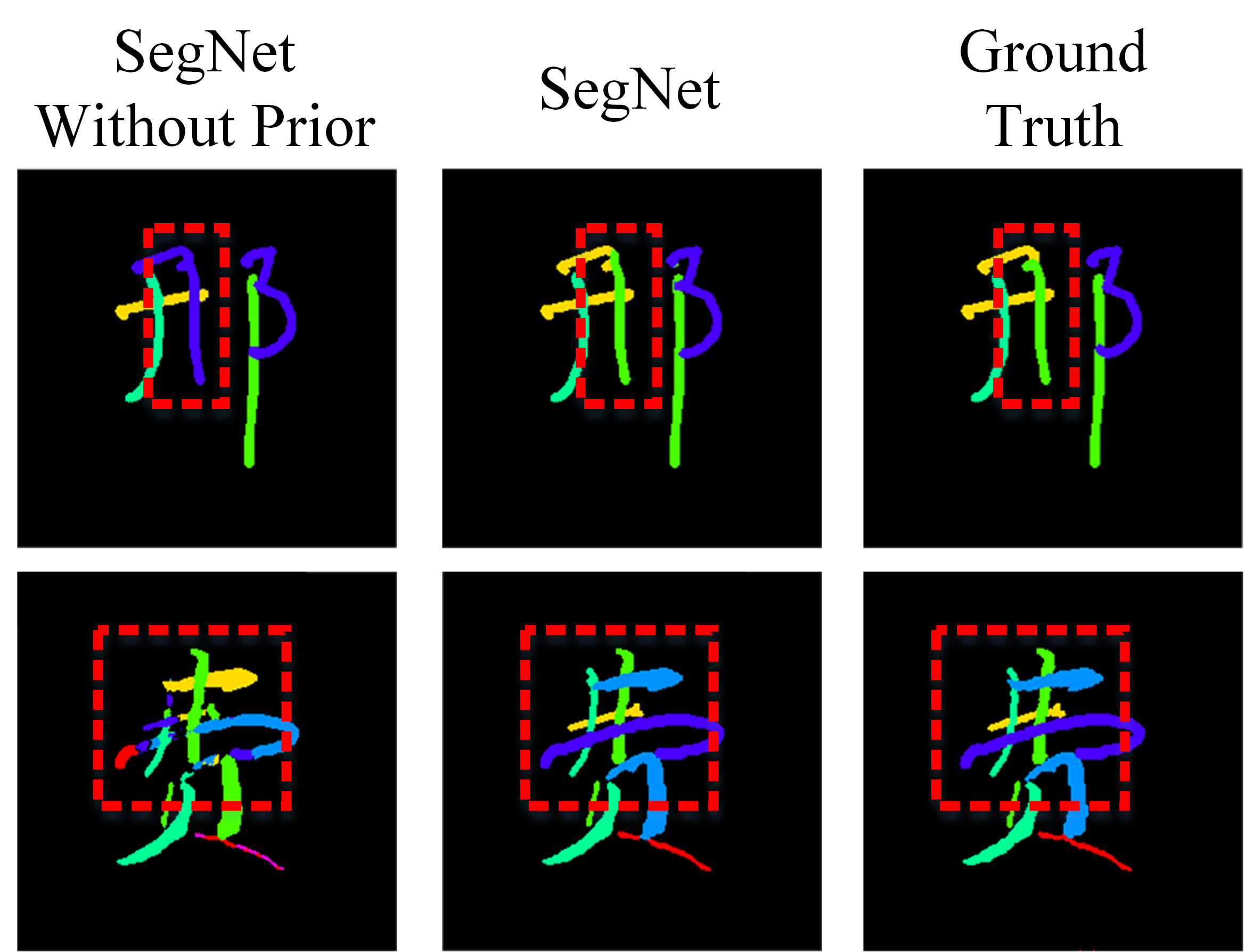} % Reduce the figure size so that it is slightly narrower than the column.
    \caption{The effect of prior information on semantic segmentation results in the SegNet.}
    \label{fig6}
\end{figure}

\begin{table}[th]
    \centering
    \normalsize
    \resizebox{1.0\columnwidth}{!}{
    \begin{tabular}{c|c|c}
        \hline   % table title
        \multirow{2}{*}{}      & \multicolumn{2}{|c}{$mIOU$ of Semantic Segment}  \\
                         \cline{2-3}
                          & CCSEDB   &     RHSEDB     \\
        \hline
        \textbf{SegNet}	&\textbf{0.917}	&\textbf{0.908}	 \\
        \hline
        SegNet Without Prior	&0.735	&0.768	\\
        \hline
    \end{tabular}}
    \caption{Effect of prior information in the SegNet.}
    \label{table4}
\end{table}

As shown in Table 4 and Figure 6, the prior information has a significant effect on the SegNet because of the high similarity between strokes. 
Compared to the SegNet, the prior information has a greater effect on the ExtractNet, as shown in Table 5. 
This is because that the prior information provides major information for analyzing the position and shape of the current single stroke in the ExtractNet. 
Semantic information has a few effect on the ExtractNet. This is because the Target Data can supplement more information when the Segment Data lacks. 
Using semantic information can further improves the precision of ambiguous strokes, as shown in Figure 7.

\begin{figure}[ht]
    \centering
    \includegraphics[width=1.0\columnwidth]{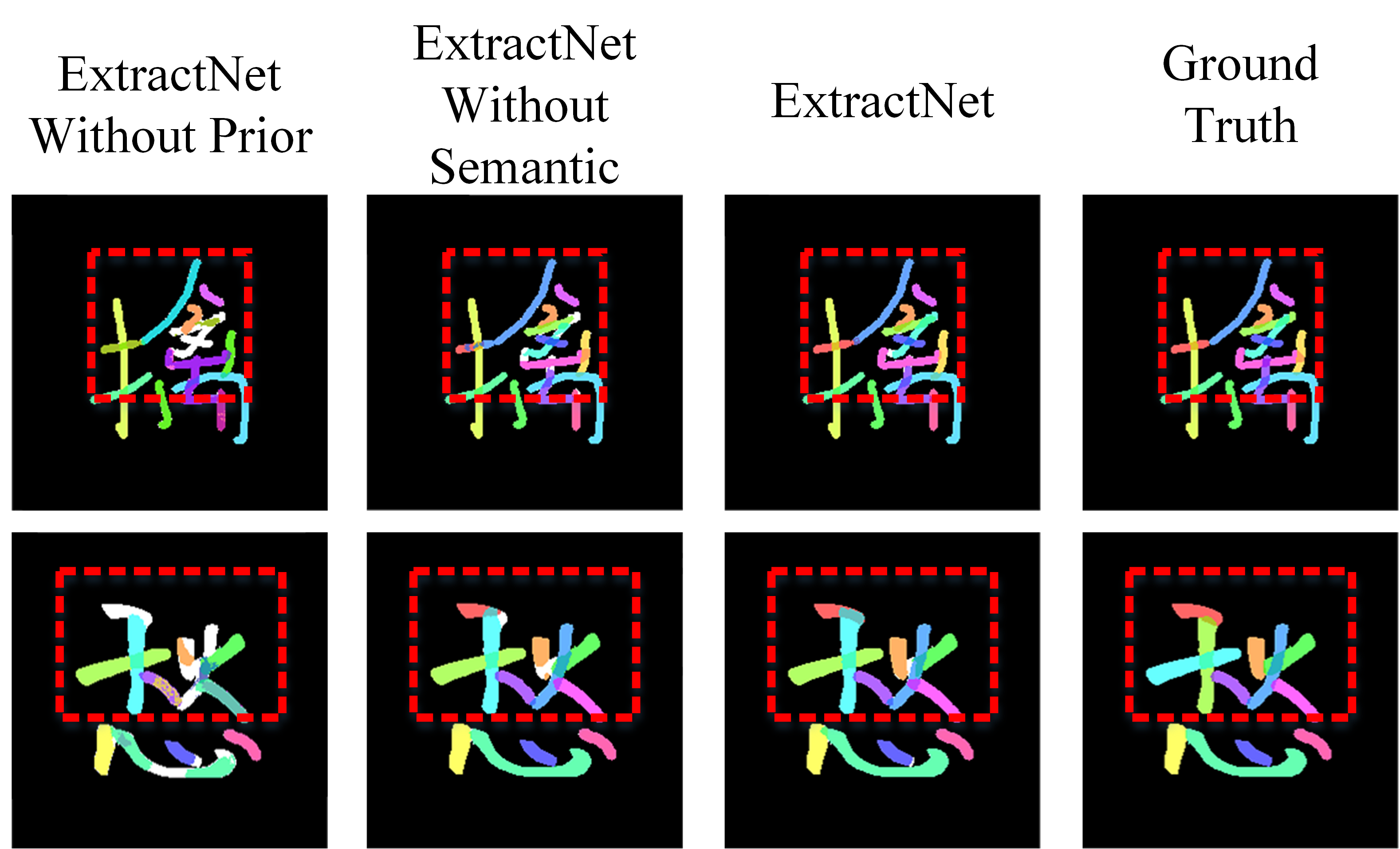} % Reduce the figure size so that it is slightly narrower than the column.
    \caption{Effect of prior and semantic information on extraction result in the ExtractNet.}
    \label{fig7}
\end{figure}

\begin{table}[ht]
    \centering
    \resizebox{1.0\columnwidth}{!}{
    \begin{tabular}{c|c|c}
        \hline   % table title
        \multirow{2}{*}{}      & \multicolumn{2}{|c}{$mIOU_m$}  \\
                         \cline{2-3}
                          & CCSEDB   &     RHSEDB     \\
        \hline
        \textbf{ExtractNet}	&\textbf{0.924}	&\textbf{0.917}	 \\
        \hline
        ExtractNet Without prior	&0.567	&0.556	\\
        \hline
        ExtractNet Without Semantic	&0.914	&0.890	\\
        \hline
    \end{tabular}}
    \caption{Effect of prior and semantic information in the ExtractNet.}
    \label{table5}
\end{table}

\subsection{Limitations}
Finally, as shown in Figure 8, we show typical errors in the results of our method.
\begin{figure}[t]
    \centering
    \includegraphics[width=1.0\columnwidth]{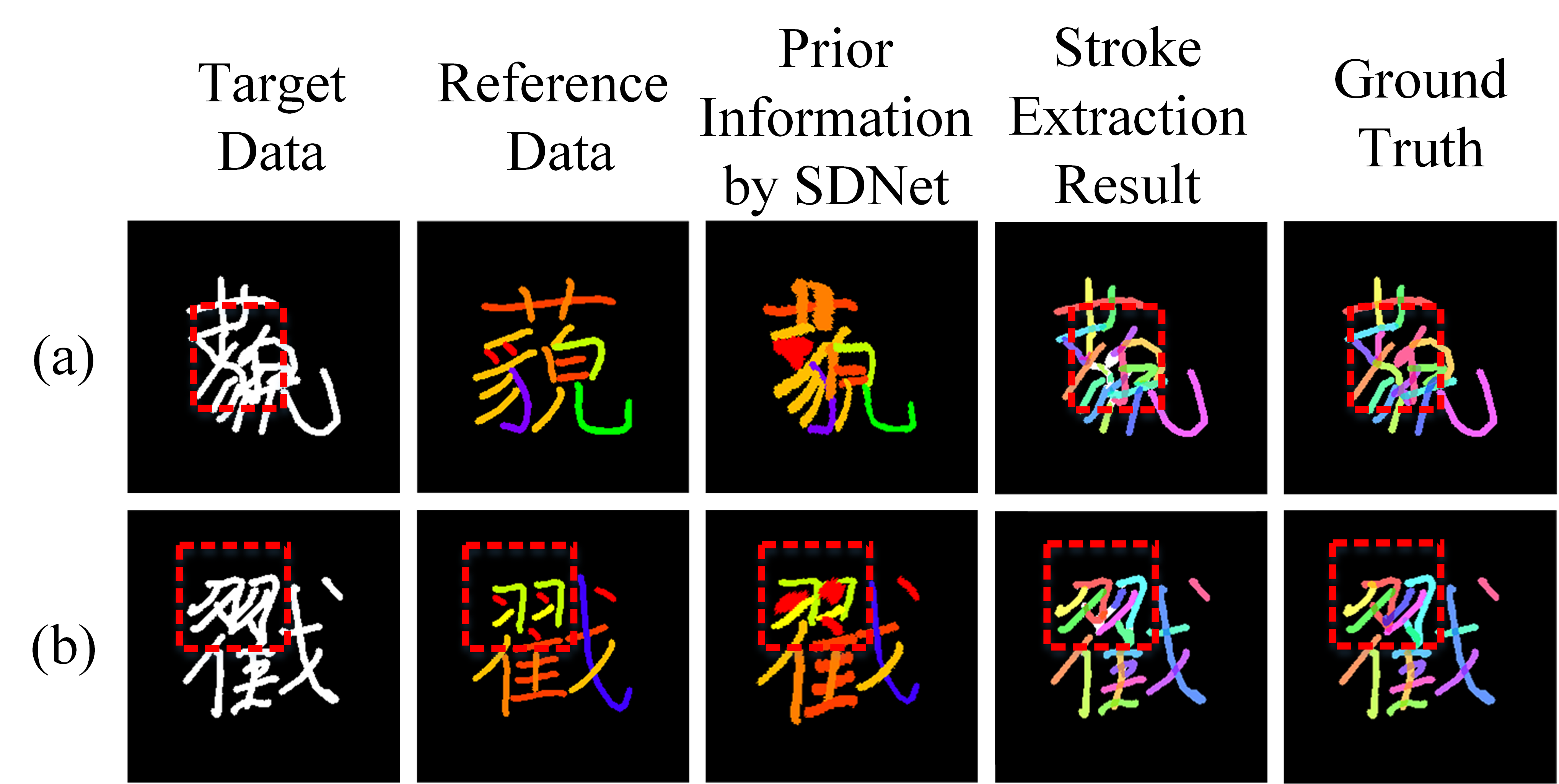} % Reduce the figure size so that it is slightly narrower than the column.
    \caption{Typical error examples of our method.}
    \label{fig8}
\end{figure}
\begin{itemize}
    \item \textbf{Scribbled strokes:} Scribbled strokes often lead to densely intersected strokes and indistinguishable stroke morphology, 
    which usually lead to failure of stroke segmentation and single-stroke extraction, such as (a) in Figure 8. 
    \item \textbf{Excessive difference in local stroke structure:} As shown in Figure 8 (b), excessive local structure differences between reference and target 
    usually lead to large registration errors, which makes the prior information of these local strokes unusable.
\end{itemize}

\section{Conclusion}
In this paper, we propose an efficient stroke extraction model of Chinese characters which takes semantic features and prior information of strokes into consideration 
efficiently. In our method, SDNet establishes the registration relationship between reference strokes and target strokes to provide the prior information of stroke 
position and shape to the target character. The prior information can guide SegNet to segment roughly the strokes of the target character. 
With the prior information and segmentation results, every stroke is extracted high precision through ExtractNet. 
Furthermore, to solve the registration problem of characters with complex stroke structures, we propose a new method for Chinese character image registration called SDNet. 
The use of the local linear stroke spatial transformation method in SDNet ensures deformability of stroke structure while maintaining the stability 
of single stroke shape in transformation.

Experiments show that our method performs better in stroke extraction than baseline methods and can be used for a wide range of characters including calligraphic 
characters and handwritten characters. And, SDNet performs better in the registration of image structure. 

We believe that prior information and stroke semantic information are the keys to the stroke extraction of Chinese characters. 
In our future work, we will pay more attention to studying new image registration methods based on SDNet and stroke segmentation methods for irregular characters.

\bibliography{aaai23}

\end{document}